\documentclass[letterpaper]{article} 
\usepackage{aaai2027}  
\nocopyright 
\usepackage[hyphens]{url}  
\usepackage{graphicx} 
\usepackage{natbib}  
\usepackage{caption} 
\usepackage{algorithm}
\usepackage{algorithmic}
\usepackage{amsmath}
\usepackage{amssymb}

\usepackage{newfloat}
\usepackage{listings}
\DeclareCaptionStyle{ruled}{labelfont=normalfont,labelsep=colon,strut=off} 
\floatstyle{ruled}
\newfloat{listing}{tb}{lst}{}
\floatname{listing}{Listing}

\usepackage{booktabs}
\usepackage{multirow}
\usepackage[table]{xcolor}
\usepackage{makecell}

\title{Practical Noise Modeling for SPAD Intensity Imaging}
\author{
    Wendi Liu,
    Yujie Lu,
    Zengxi Zhang,
    Haiyang Jiang,
    Weihang Ran,
    Yinqiang Zheng\corresponding
}
\affiliations{
    The University of Tokyo\\

}

\begin{document}

\maketitle

\begin{abstract}
  Single-photon avalanche diode (SPAD) cameras are promising for low-light and high-dynamic-range intensity imaging, but their practical use is limited by complex sensor-specific noise. Unlike time-correlated single-photon counting (TCSPC) systems, SPAD cameras record whether at least one detection occurred in each gate without photon timestamps in intensity imaging mode, making explicit noise decomposition difficult. We present a practical noise modeling and calibration framework for SPAD intensity denoising. Our forward model describes binary-frame accumulation with a Binomial observation process, models signal-independent dark noise as an exposure-dependent pure dark count term plus an exposure-independent dark-frame bias term, and incorporates pixel-wise response non-uniformity. We design a dedicated calibration procedure for the proposed model and use it to build a count-domain noise-synthesis pipeline for network training. For denoising, we further design a SPAD-specific dark-shading correction (SPAD-DSC) to remove most systematic noise before network training. We construct a real-world SPAD intensity dataset for testing. Experimental results demonstrate the superiority of the proposed noise model.
\end{abstract}


\section{Introduction}
\label{sec:intro}
\begin{figure}[h]
    \centering
    \includegraphics[width=0.85\columnwidth]{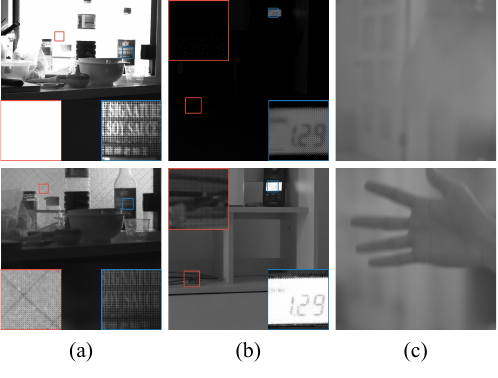}
    \caption{CMOS RAW (top) vs. SPAD RAW (bottom) in (a) HDR, (b) low-light (1.29 lx), and (c) low-light high-speed scenes. (a) SPAD RAW preserves both highlight and shadow details in a single shot and (b) captures information in photon-starved environments with high sensitivity, (c) whereas CMOS RAW requires longer exposure to collect sufficient signal under the same low-light condition and therefore suffers from motion blur in high-speed scenes.}
    \label{fig:cvs}
\end{figure}

Single-photon avalanche diodes (SPADs) combine single-photon sensitivity with picosecond-scale timing resolution, and have enabled a broad range of scientific and industrial applications, including LiDAR \cite{SPADLiDAR}, fluorescence lifetime imaging \cite{nedbal2024time}, and quantum communication \cite{buller2009single,SPADApplications}. Recent advances in large-format SPAD arrays \cite{CanonMS500,MegapixelSPAD,3.2MegapixelSPAD,SwissSPAD} are extending this capability from specialized time-resolved sensing toward general-purpose intensity imaging. Compared with conventional CMOS cameras, SPAD sensors directly count photon-triggered avalanches and introduce negligible readout noise. Their high sensitivity is particularly advantageous in photon-starved environments, while flexible gating and frame accumulation provide the potential for wide-dynamic-range acquisition. These properties make SPADs a promising imaging modality for low-light and high-dynamic-range (HDR) photography \cite{QBP,photoninhibition,PanoramasFromPhotons,ingle2021passive}, as illustrated in Fig.~\ref{fig:cvs}.

Despite these advantages, SPAD intensity images are affected by complex sensor-specific non-idealities, including dark counts, pile-up, dead time, pixel response non-uniformity, crosstalk, and afterpulsing. These effects are coupled and vary across pixels and operating conditions, making SPAD noise difficult to characterize accurately or model explicitly.

Many existing SPAD noise models are developed for time-correlated single-photon counting (TCSPC) or other time-resolved applications. TCSPC records photon arrival times relative to a synchronization signal, whereas intensity imaging uses fixed-duration exposure gates and records only whether at least one detection occurred in each gate. Photon timestamps are therefore unavailable in the intensity-imaging setting, and commercial SPAD cameras designed for intensity and video imaging commonly expose accumulated image measurements rather than photon timestamps \cite{CanonMS500}. Consequently, precise temporal modeling of dead time and afterpulsing, as well as causal identification of event sources, is not supported by the available observation. Conventional CMOS noise models are likewise unsuitable because their noise characteristics and approximately linear analog acquisition mechanism differ fundamentally from binary SPAD imaging. Unlike TCSPC-based models, the most relevant intensity-imaging work by Bian \textit{et al.} \cite{bian2023highresolution} attempts to explicitly separate pure dark counts, afterpulsing, and crosstalk from binary-frame sequences and then synthesize training data. Without photon timestamps or causal event labels, however, independent events can satisfy the heuristic rules of the noise model, while true correlated events may be missed, making the explicit noise decomposition ambiguous and error-prone.

In this work, we focus on noise modeling for SPAD intensity imaging under fixed-duration binary-frame accumulation, rather than TCSPC mode. First, we model the signal-independent noise observed in dark frames with a pixel-wise linear intensity model. Its exposure-dependent coefficient describes the pure dark count rate (DCR), while an exposure-independent term collectively describes the remaining dark-frame noise characteristics, including crosstalk, afterpulsing, random telegraph signal (RTS), and other complex effects that are difficult to separate explicitly. We estimate both parameters directly from multi-exposure dark-frame trigger counts using constrained Binomial maximum likelihood. Second, we model spatial response non-uniformity and calibrate a pixel-wise relative gain map over the full sensor using flat-field measurements. Finally, based on the calibrated noise model, we construct a count-domain noise-synthesis pipeline to provide diverse training data, and derive a SPAD-specific dark-shading correction (DSC) that removes most systematic noise before network training. Together, these components form a unified calibration, synthesis, correction, and learning pipeline for practical SPAD intensity denoising.

Our main contributions are three-fold:
\begin{itemize}
\item We establish a noise model for SPAD intensity imaging that describes the noise characteristics directly from observable measurements, without relying on complex explicit noise decomposition.
\item We build a closed loop of noise modeling, calibration, data synthesis, and systematic pre-correction (SPAD-DSC), where the noise parameters are calibrated according to the proposed model and guide both training-data synthesis and the pre-correction processing.
\item We construct a real-world dataset for SPAD image denoising, and experiments on it demonstrate the effectiveness of the proposed method.
\end{itemize}

\section{Related Work}
\subsection{SPAD Noise Modeling and Calibration}
SPAD sensors have been extensively studied from device, circuit, and time-resolved measurement perspectives. Prior work has identified dark count rate (DCR), photon detection efficiency variation, dead time, afterpulsing, crosstalk, and pile-up as key factors that shape SPAD measurement statistics \cite{SPADApplications,bruschini2019spadbiophotonics,dutton2016single,nonuniformityanalysis,hernandez2017computational,albeck2025dead,kang2003afterpulsing,ziarkash2018comparative,rech2008crosstalk,patting2018pileup,antolovic2018dynamic}. These studies provide important physical and statistical understanding of SPAD sensors, but they are mainly designed for device characterization, circuit analysis, or time-resolved applications such as TCSPC and time-of-flight (ToF) sensing. Their assumptions do not directly match SPAD intensity imaging, where the available RAW observation is an accumulated count image generated from fixed-duration binary gates.

For SPAD intensity denoising, the most relevant work is Bian \textit{et al.} \cite{bian2023highresolution}, which synthesizes training data by explicitly decomposing dark observations into pure dark counts, afterpulsing, and crosstalk, providing an important step toward physics-informed SPAD image restoration. However, as discussed in the introduction, such explicit decomposition is inherently ambiguous without photon timestamps or causal event labels. This reflects a broader difficulty of SPAD intensity imaging: multiple sensor non-idealities are coupled in the accumulated count observation, making exact decomposition and explicit noise modeling difficult.

\subsection{Noise Modeling for CMOS}
Noise modeling for conventional CMOS cameras is commonly studied in the RAW domain, where the photo-response is approximately linear and noise is often modeled as signal-dependent shot noise plus signal-independent electronics-induced components \cite{foi2008poissongaussian,wang2020practical}. Modern denoising algorithms often learn the mapping from paired short/long exposure RAW data \cite{chen2018learningtosee,Chen_2019_ICCV,CRVD}. To reduce the cost of paired data collection, physics-based methods synthesize training pairs from calibrated noise models by explicitly separating noise sources \cite{wei2021physicsbased,PMN,EMCCD,MSFA,PolarizedDenoising,LED}, sampling real noise \cite{SFRN,NMOH,SWIR}, or learning noise distributions from noisy observations \cite{PNNP,LRD,LLD,dancingunderthestars,noiseflow}. Recent minimal-acquisition methods further reduce calibration effort; for example, 2-Shots in the Dark models signal-dependent noise with a Poisson process and synthesizes signal-independent noise from limited dark-frame data using Fourier-domain spectral sampling \cite{lu2026twoshots}. These works demonstrate the value of accurate sensor modeling, calibration, and dark-shading correction for RAW denoising. However, their underlying imaging mechanism does not match SPAD intensity imaging, which motivates SPAD-specific noise modeling research.

\section{Noise Model}
\subsection{Model Formulation}
\subsubsection{Binary-Frame Accumulation.}
We consider the SPAD intensity-imaging mode in which each pixel records a binary value during a fixed exposure gate. Let $\lambda\geq0$ denote the expected number of effective events within one gate. Assuming Poisson event arrivals, the probability that the pixel is triggered at least once is
\begin{equation}
    p=1-\exp(-\lambda).
    \label{eq:spad-trigger}
\end{equation}
The binary observation is therefore $B_n\sim\operatorname{Bernoulli}(p)$. An intensity RAW image accumulates $N$ such binary frames, following the Binomial distribution with respect to $N$ and $p$:
\begin{equation}
    X=\sum_{n=1}^{N}B_n\sim\mathcal B(N,p).
    \label{eq:spad-accumulation}
\end{equation}
Here $X\in\{0,\ldots,N\}$ is the accumulated count available to the imaging pipeline. Eq.~\ref{eq:spad-trigger} explicitly describes the pile-up effect of the SPAD camera: multiple events occurring within the same gate still produce only one binary trigger, leading to nonlinear response while the illuminance or exposure time changes linearly.

\subsubsection{Signal-Independent Dark Noise.}
With incident light blocked, dark frames still contain detections from multiple sensor noise sources. The major component follows Poisson arrival statistics with a stable pixel-dependent rate, producing a repeatable spatial pattern analogous to fixed-pattern noise in CMOS sensors. We model this component as pure dark counts and denote its rate, i.e., the dark count rate (DCR), by $D_k$. The remaining dark-frame detections contain crosstalk, afterpulsing, random telegraph signal (RTS), and other complex effects that are difficult to separate explicitly; we collectively model their aggregate contribution as an exposure-independent expected count $D_b$ per binary gate. The total expected dark-event count in one gate is
\begin{equation}
    \lambda_{\mathrm{dark}}(\Delta t)=D_k \Delta t+D_b.
    \label{eq:dark-intensity}
\end{equation}

\subsubsection{Spatial Non-Uniformity and Full Forward Model.}
\begin{figure}[h]
    \centering
    \includegraphics[width=\columnwidth]{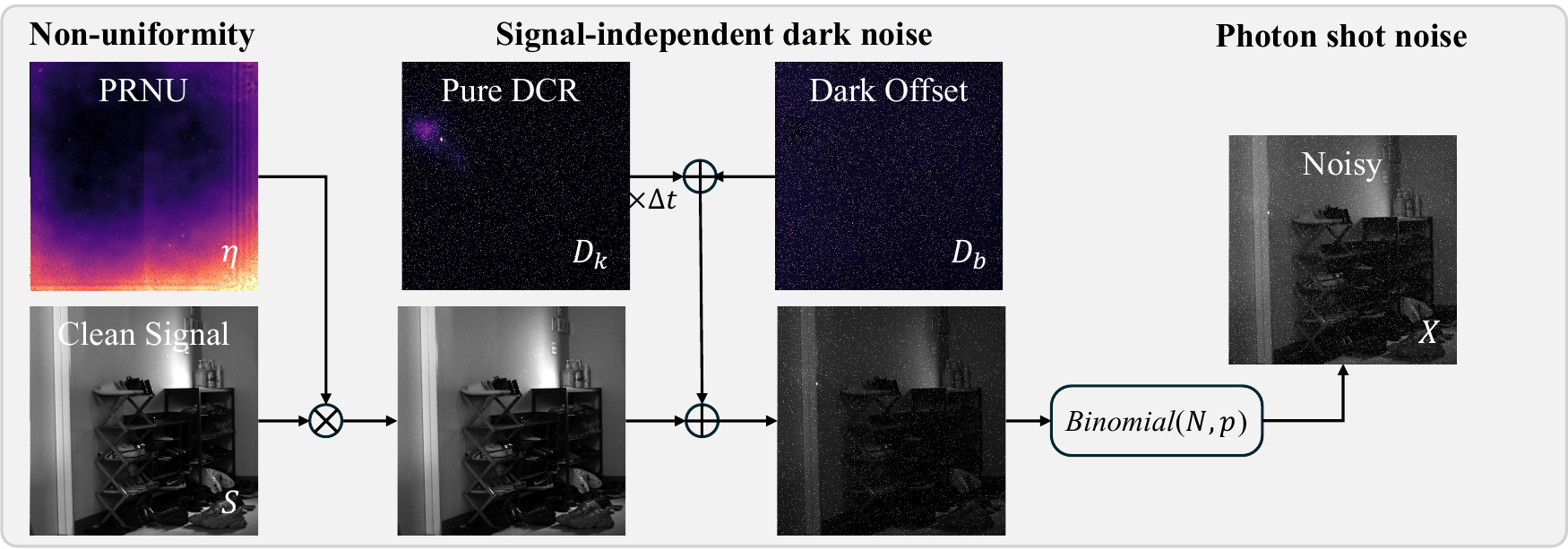}
    \caption{Overview of SPAD intensity imaging pipeline and visualization of noise components and signal response at different stages. }
    \label{fig:forward}
\end{figure}

Fabrication variations cause the quantum efficiency and other device responses to differ across pixels, resulting in a spatially varying photon detection efficiency (PDE). Consequently, the same incident illumination produces different expected signal intensities at different pixel locations. We model this pixel response non-uniformity (PRNU) by a pixel-wise relative response map $\eta$, normalized with respect to a reference response. Let $S\geq0$ denote the clean signal intensity accumulated over $N$ gates at the reference response. The expected signal at each pixel is then degraded to $\eta S$ due to spatial response non-uniformity. Combining this signal response with Eqs.~\ref{eq:spad-trigger}--\ref{eq:dark-intensity} gives the forward degradation model from $S$ to the accumulated measurement $X$:
\begin{equation}
    X\sim\mathcal B\left(
    N,\,
    1-\exp\left[
    -\frac{\eta S}{N}
    -D_k\Delta t
    -D_b
    \right]\right).
    \label{eq:full-forward-model}
\end{equation}
The whole process is shown in Fig.~\ref{fig:forward}. All operations in Eq.~\ref{eq:full-forward-model} are applied pixel-wise. 
\subsection{Parameter Calibration}
\label{sec:calib}
For calibration and denoising experiments, we use the SPAD Alpha camera from PI Imaging \cite{PIImagingDocs}. It supports both high-speed 1-bit burst acquisition and accumulated intensity imaging up to 12-bit, allowing us to estimate binary-frame trigger statistics and evaluate the same model in the accumulated RAW domain, and its $1024\times1024$ resolution is sufficient for network training and practical scene capture while remaining tractable for exhaustive pixel-wise calibration.

We collected three types of calibration data under the target camera configuration: bias frames to calibrate $D_k$ and $D_b$, flat-field frames to calibrate the relative gain map $G=1/\eta$, and a ColorChecker sequence to calibrate color-visualization parameters. 
\subsubsection{Bias Frame Calibration.}
Bias frames are captured with incident light blocked so that the observed triggers are entirely sensor-induced. We capture binary dark-frame bursts at 10 exposure settings from $1~\mu s$ to $1200~\mu s$, with $2\times10^4$ to $10^5$ valid frames per setting. For each pixel and exposure $t_i$, we count the number of triggered binary frames $k_i$ among $N_i$ valid observations and fit the multi-exposure trigger statistics directly in the Binomial domain:
\begin{equation}
    k_i \sim \mathcal B\left(
    N_i,\; 1-\exp\left[-(D_k t_i + D_b)\right]
    \right).
    \label{eq:dkdb_mle}
\end{equation}
We constrain $D_k\ge 0$ and $D_b\ge 0$ to ensure the physical validity of the model, and estimate the two parameters by constrained Binomial maximum likelihood estimation (MLE):
\begin{equation}
    (\hat D_k,\hat D_b)
    =\arg\min_{\substack{D_k\ge 0\\ D_b\ge 0}}
    \mathcal{L}(D_k,D_b),
    \label{eq:dkdb_objective}
\end{equation}
\begin{equation}
    \begin{aligned}
    \mathcal{L}(D_k,D_b)
    =\sum_i\Big[
    &-k_i \log\!\big(1-\exp[-(D_k t_i+D_b)]\big) \\
    &+(N_i-k_i)(D_k t_i+D_b)
    \Big].
    \end{aligned}
    \label{eq:dkdb_loss}
\end{equation}
We optimize Eq.~\ref{eq:dkdb_objective} using projected damped Newton updates.

The same multi-exposure calibration is also used to mark bad pixels. We mark pixels with abnormally high DCR or intercept values, as well as pixels whose constrained MLE fails to converge, as bad pixels; detailed thresholds and category definitions are provided in the supplementary material. We save the pixel-wise $D_k$ and $D_b$ maps together with the bad-pixel mask for the subsequent noise synthesis and denoising pipeline. 

\subsubsection{Flat-field Frames Calibration.}

Flat-field calibration is performed from 1-bit sequences captured with spatially uniform illumination. We sweep $\Delta t$ from $2~\mu s$ to $50~\mu s$ so that the measurements cover a broad dynamic range of flat-field responses. For every $\Delta t$, we accumulate the 1-bit frames into multiple 4-, 6-, 7-, 8-, 10-, and 12-bit count images and average them to reduce the impact of shot noise, then retain the highest bit-depth unsaturated setting for gain calibration. For a selected flat-field count image $C$ accumulated from $N$ binary frames with per-frame exposure $\Delta t$, we first compute
\begin{equation}
    p=\frac{C}{N},
\end{equation}
and then map the flat-field counts $C$ back to the linear domain by pile-up inversion, and subtract the previously calibrated dark-noise terms to obtain a clean linear response estimate:
\begin{equation}
    R=\frac{-\ln(1-p)-D_b}{\Delta t}-D_k.
    \label{eq:flatfield_response}
\end{equation}
For each color filter array (CFA) channel $c$, let $R_{c}^{\mathrm{ref}}$ denote the global median response of that channel over the flat-field image. We then define
\begin{equation}
    G_c(x)=\frac{R_{c}^{\mathrm{ref}}}{R_c(x)}.
\end{equation}
In this way, we obtain one gain map for each CFA channel. We mark pixels whose response is below $20\%$ of their local neighborhood response as dead pixels, and merge this dead-pixel mask with the dark-frame bad-pixel mask for the subsequent synthesis and denoising pipeline.

\subsection{Synthesis}
Large-scale paired clean/noisy SPAD RAW data are difficult to capture in practice, because obtaining a noise-free reference under identical scene, optics, and sensor conditions is costly and often infeasible. Following the common strategy in CMOS RAW denoising of training on calibrated synthetic pairs \cite{PMN,SFRN,wei2021physicsbased}, we construct synthetic SPAD training data from clean linear RAW images.

We use the See-in-the-Dark (SID) dataset \cite{chen2018learningtosee}, as it provides clean long-exposure linear RAW images with low residual noise. Let $\mathbf{I}_{\mathrm{SID}}$ denote a clean SID RAW image in the original 14-bit CMOS domain. For a target SPAD bit depth $b$, the accumulation count $N$ is fixed by the sensor setting. We first linearly map $\mathbf{I}_{\mathrm{SID}}$ to the corresponding SPAD reference range and use the mapped image $\mathbf{I}^{(b)}$ as the synthetic clean target in that bit-depth domain. Here the reference range denotes the count upper bound in the SPAD linear domain for the corresponding bit-depth setting, which is derived from the pile-up inverse response $-N\ln(1-p)$ with $p=C/N$. We use the reference values provided by PI Imaging \cite{PIImagingDocs} to replace the infinite upper bound when $p\rightarrow 1$. The corresponding black level is mapped in the same way and denoted by $B^{(b)}$.

For each pixel, the signal term used in synthesis is obtained by removing the mapped black level:
\begin{equation}
    S=\max\left(I^{(b)}-B^{(b)},\,0\right).
\end{equation}
We then synthesize noisy SPAD counts using the forward model in Eq.~\ref{eq:full-forward-model}. Since $N$ is determined once the target bit depth is fixed, we vary the dark-noise exposure time $T_{\mathrm{dcr}}$, or equivalently the per-frame exposure $\Delta t=T_{\mathrm{dcr}}/N$, to generate different noise levels from the same clean RAW source. In our experiments, we use $T_{\mathrm{dcr}}\in\{30,60\}\,\mathrm{ms}$ together with 8-bit and 12-bit settings, yielding four synthesis configurations for training.

\subsection{SPAD-DSC}
In CMOS RAW denoising, dark-shading correction is often applied before network training to remove systematic sensor noise, so that the network mainly handles the remaining stochastic shot noise \cite{PMN,NMOH,EMCCD}. This strategy has been widely used in calibrated RAW pipelines and has shown strong empirical performance. Inspired by this, we design a SPAD-specific dark-shading correction (SPAD-DSC) to remove most systematic noise and reduce the burden of learning complex sensor-specific artifacts.

Given an accumulated SPAD count $X$ from $N$ binary frames and total exposure time $T$, SPAD-DSC first maps the count back to the linear domain by pile-up inversion, then subtracts the calibrated dark component $D_kT+ND_b$, and compensates the response non-uniformity with the gain map $G$:
\begin{equation}
    \hat S=\left[-N\ln\left(1-\frac{X}{N}\right)-\left(D_kT+ND_b\right)\right]\times G.
\label{eq:spad-dsc}
\end{equation}
For saturated pixels with $X\approx N$, we use the same saturation handling as in the synthesis pipeline. Finally, using the previously calibrated bad-pixel mask, we repair bad-pixel locations by neighborhood-mean inpainting to avoid numerical instability when applying Eq.~\ref{eq:spad-dsc}.

\subsection{Color ISP}
For color imaging, we additionally calibrate white-balance (WB) gains and a color correction matrix (CCM) using ColorChecker captures. These parameters are used in our color ISP pipeline to convert the RAW measurements into visually meaningful color images for display and evaluation. More details are provided in the supplementary material.

\section{Experiments}
\subsection{Settings}
\begin{figure*}[t]
    \centering
    \includegraphics[width=0.95\textwidth]{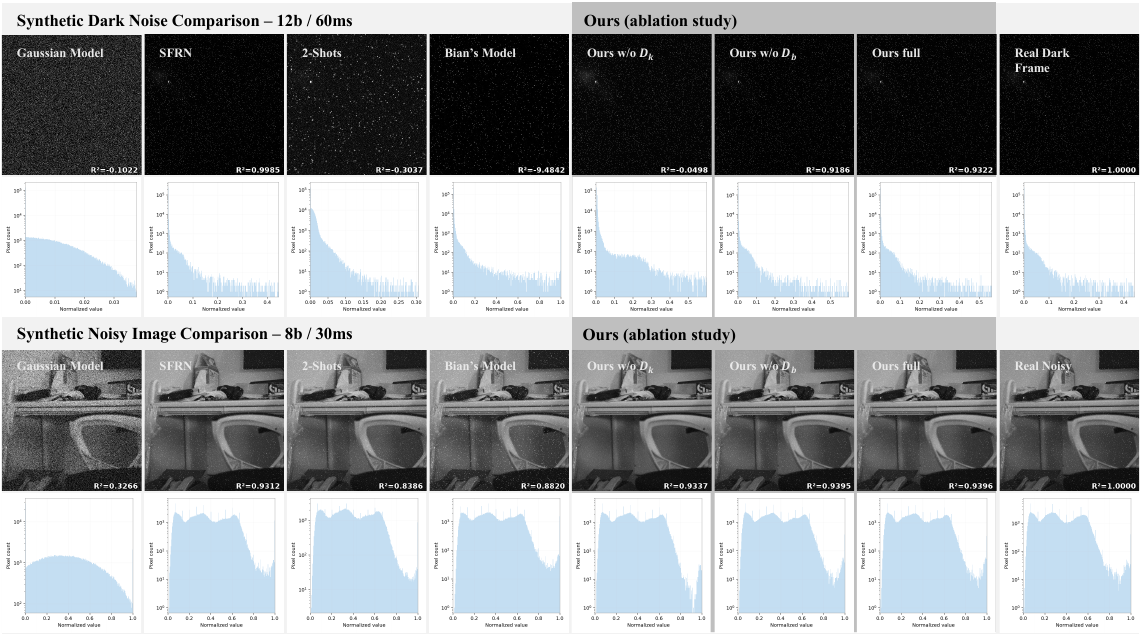}
    \caption{Visualization of noise synthesis. The upper part compares synthetic dark frames under the 12b/60ms setting, and the lower part compares synthetic noisy RAW images under the 8b/30ms setting.}
    \label{fig:noise_synthesis}
\end{figure*}
\subsubsection{Training.} We train the denoising network on synthetic SPAD data generated from the SID dataset \cite{chen2018learningtosee}. We pack each $2848\times4256$ SID RAW image into four Bayer channels and crop $512\times512\times4$ patches as clean training targets. Using the calibrated forward model in Eq.~\ref{eq:full-forward-model}, we synthesize noisy SPAD measurements under two representative accumulation modes: 8-bit imaging with $N=255$ binary frames and 12-bit imaging with $N=4080$ binary frames\footnote{A 12-bit image is obtained by accumulating 16 8-bit images \cite{PIImagingDocs}; therefore, $N=16\times255=4080$.}. For each bit depth, we set the equivalent dark-noise exposure time $T_{\mathrm{dcr}}$ to $30$ ms and $60$ ms. The network is trained on a mixture of these four synthesis settings and directly tested on inputs of different bit depths and exposure times without exposure-specific fine-tuning. We report two variants of our method: one directly trains the network on the synthesized paired clean/noisy data, denoted by Ours in Tab.~\ref{tab:noise_model_comparison}, and the other applies the proposed SPAD-DSC to remove most systematic noise before network training, denoted by Ours* in Tab.~\ref{tab:noise_model_comparison} and Tab.~\ref{tab:ablation}.

\subsubsection{Evaluation.} 
\begin{figure}[h]
    \centering
    \includegraphics[width=\columnwidth]{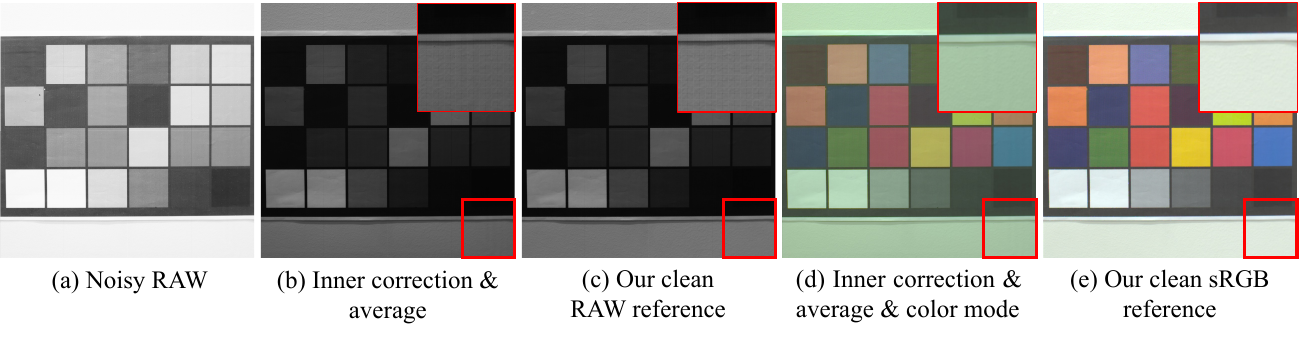}
    \caption{Construction of the clean reference in our real-world test set. The camera's internal correction still leaves vertical stripes and lens vignetting caused by spatially non-uniform response, as well as color cast in the color mode. With our recalibrated gain and color parameters, we obtain visually better references.}
    \label{fig:sample}
\end{figure}

We evaluate denoising performance on a real SPAD intensity test set captured with the SPAD Alpha camera. For each scene, we first enable the camera's internal pile-up correction and noise correction, capture and average about 100 frames under the same camera setting, and then apply non-uniformity correction and bad-pixel correction to obtain the clean reference image, as shown in Fig.~\ref{fig:sample}. The exposure time is selected according to scene brightness and ranges from 5 ms to 200 ms, with either 8-bit or 12-bit accumulation. We then disable the camera's internal corrections and capture one noisy image under the same scene, exposure, and bit-depth setting. In total, we collect 100 paired noisy-clean real samples. For color evaluation, all methods use the same calibrated WB gains and CCM for sRGB rendering. We report PSNR, SSIM, and LPIPS in both the RAW domain and the sRGB domain.

\subsubsection{Comparison Methods.} We compare our method with four noise-model baselines: Gaussian noise model \cite{foi2008poissongaussian}, SFRN \cite{SFRN}, Bian's model \cite{bian2023highresolution}, and 2-Shots \cite{lu2026twoshots}. Since SFRN and 2-Shots are originally designed for CMOS RAW noise modeling, we adapt them to the SPAD setting by modeling the signal shot noise with the binary-frame accumulation model in Eq.~\ref{eq:spad-accumulation}. We capture matched dark frames for each exposure time and bit-depth setting for SFRN and 2-Shots, and recalibrate Bian's model following the original procedure; detailed formulations are provided in the supplementary material. SPAD-DSC is a systematic-noise correction procedure derived from our noise model. Since Gaussian, SFRN, and Bian's model do not model the corresponding SPAD noise components, SPAD-DSC cannot be applied to them, and these baselines are trained end-to-end using their own synthesized noisy/clean pairs. The 2-Shots method includes a dark-shading correction step, so we follow its original setting by capturing matched dark frames and applying its own DSC before training. All methods train the same denoising backbone and are tested on the same real test set.

\begin{figure*}[t]
    \centering
    \includegraphics[width=0.95\textwidth]{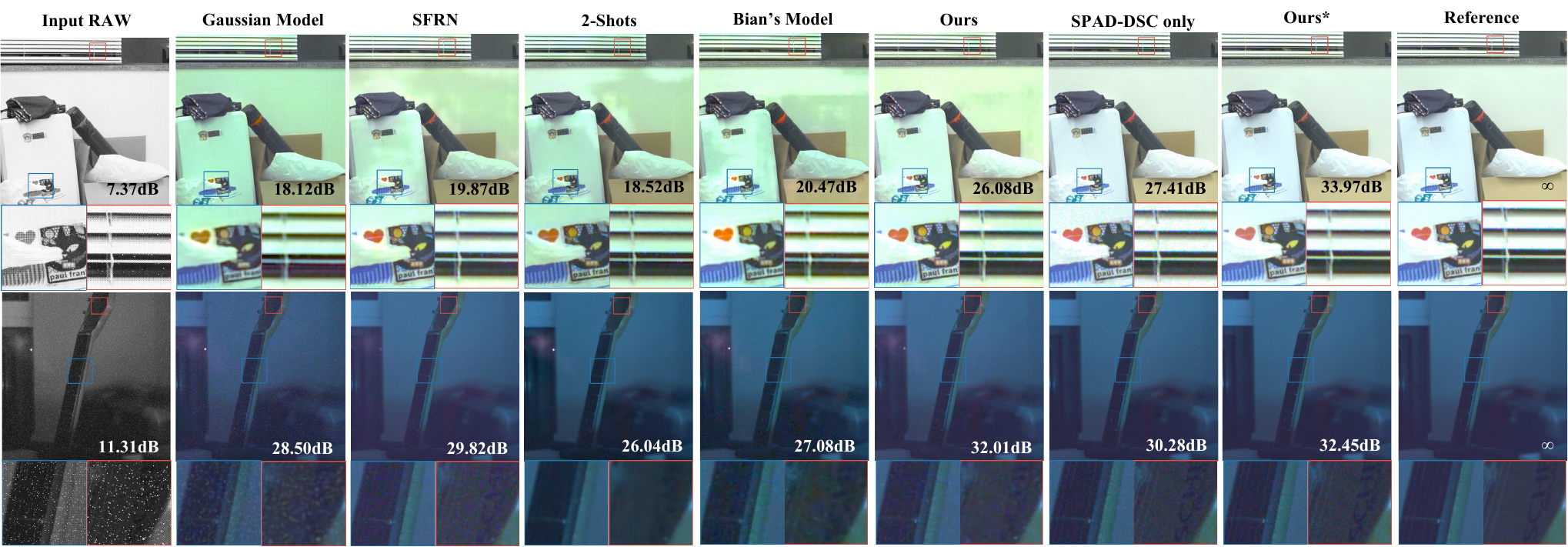}
    \caption{Comparison of denoising performance on real-world SPAD scenes. PSNR is reported for each result, and the blue and red boxes mark the zoomed-in regions shown below each image. Baseline noise models leave residual noise, color artifacts, or degraded local structures, whereas Ours suppresses noise more thoroughly and Ours* further removes the remaining systematic artifacts, producing results closest to the reference.}
    \label{fig:real_world_comparison}
\end{figure*}

\subsection{Noise Model Comparison}
We first evaluate noise synthesis under four imaging configurations, combining 8-bit and 12-bit accumulation with exposure times of 30 ms and 60 ms. For dark-frame evaluation, we capture 10 real dark frames for each configuration and randomly select one as the real reference. SFRN directly samples dark noise from the remaining captured frames except for the real reference, whereas 2-Shots uses one of these frames to estimate its method-specific noise parameters. We report the coefficient of determination ($R^2$) for dark-frame synthesis in Tab.~\ref{tab:noise_synthesis_ablation}. We further visualize both dark-frame synthesis and noisy-image synthesis in Fig.~\ref{fig:noise_synthesis}.

As shown in Tab.~\ref{tab:noise_synthesis_ablation}, SFRN naturally achieves the highest dark-frame $R^2$ scores because it directly samples noise from real dark frames. Apart from SFRN, our method performs consistently well across all four configurations, whereas the noise magnitude and spatial distributions synthesized by the other models exhibit substantial mismatches with the real measurements, resulting in negative $R^2$ scores. The visual comparisons in Fig.~\ref{fig:noise_synthesis} further show that our model better reproduces not only the dark-frame distribution but also the noise appearance and histogram of real noisy images.

\begin{table}[h]
    \centering
    \resizebox{0.47\textwidth}{!}{
    \begin{tabular}{lcccc}
    \toprule
    \textbf{Noise Model}
    & \textbf{8b/30ms}
    & \textbf{8b/60ms}
    & \textbf{12b/30ms}
    & \textbf{12b/60ms} \\
    \midrule

    Gaussian
    &-4.1183  &-2.4054  &-0.2947  &-0.1022  \\

    SFRN
    &\textbf{0.9854}  &\textbf{0.9872}  &\textbf{0.9976}  &\textbf{0.9985}  \\

    Bian's Model
    &-0.6108  &-0.1491  &-26.4845  &-9.4842  \\

    2-Shots
    &-0.3386  &-0.3497  &-0.2774  &-0.3037  \\

    \midrule

    Ours w/o $D_k$
    &0.8167  &0.6825  &-3.2282  &-0.0498  \\

    Ours w/o $D_b$
    &0.9798  &0.9729  &\underline{0.7916}  &0.9186  \\

    \midrule

    Ours
    &\underline{0.9814}  &\underline{0.9755}  &0.7870  &\underline{0.9322}  \\

    \bottomrule
    \end{tabular}
    }
    \caption{Comparison of noise synthesis fidelity. We report $R^2$ between synthesized and real dark frames under different bit-depth and exposure-time settings.}
    \label{tab:noise_synthesis_ablation}
\end{table}
\begin{table}[h]
    \centering
    \resizebox{\columnwidth}{!}{
    \begin{tabular}{llcccccc}
    \toprule
    \multirow{2}{*}{\textbf{Noise Model}} & \multirow{2}{*}{\textbf{Denoiser}} 
    & \multicolumn{3}{c}{\textbf{RAW Domain}} 
    & \multicolumn{3}{c}{\textbf{sRGB Domain}} \\
    \cmidrule(lr){3-5} \cmidrule(lr){6-8}
    & & \textbf{PSNR$\uparrow$} & \textbf{SSIM$\uparrow$} & \textbf{LPIPS$\downarrow$}
    & \textbf{PSNR$\uparrow$} & \textbf{SSIM$\uparrow$} & \textbf{LPIPS$\downarrow$} \\
    \midrule

    Input & --
    & 9.948 & 0.2053 & 0.7743 & 8.409 & 0.4487 & 0.6088 \\
    SPAD-DSC only & --
    & 41.875 & 0.9837 & 0.0928 & 34.522 & 0.7573 & 0.2621 \\

    \midrule
    Gaussian & U-Net 
    & 30.741 & 0.9380 & 0.0901 & 25.129 & 0.8580 & 0.3503 \\
    
    SFRN & U-Net 
    & 34.624 & 0.9699 & 0.0450 & 28.786 & 0.8646 & 0.1998 \\
    
    2-Shots & U-Net 
    & 33.421 & 0.9574 & 0.0460 & 26.296 & 0.8714 & 0.2270 \\
    
    Bian's Model & U-Net 
    & 33.645 & 0.9606 & 0.0593 & 26.748 & 0.8665 & 0.3106 \\
    
    \midrule
    
    \multirow{4}{*}{Ours} 
    & U-Net 
    & 36.737 & 0.9814 & 0.0292 & 31.289 & 0.9099 & 0.1758 \\
    
    & Uformer 
    & 38.859 & 0.9878 & 0.0269 & 33.152 & 0.9270 & 0.1182 \\
    
    & Restormer 
    & 37.428 & 0.9763 & 0.0568 & 32.361 & 0.8405 & 0.1821 \\
    
    & AST 
    & 36.698 & 0.9793 & 0.0386 & 31.738 & 0.8761 & 0.1523 \\
    
    \midrule
    
    \multirow{4}{*}{Ours*} 
    & U-Net 
    & \underline{46.812} & 0.9952 & \underline{0.0152} & 38.319 & 0.9274 & 0.0882 \\
    
    & Uformer 
    & 46.545 & \textbf{0.9957} & \underline{0.0152} & 38.092 & \textbf{0.9420} & \underline{0.0848} \\
    
    & Restormer 
    & \textbf{47.611} & \underline{0.9956} & \textbf{0.0127} & \textbf{39.680} & 0.9211 & \textbf{0.0733} \\
    
    & AST 
    & 46.335 & 0.9952 & 0.0173 & \underline{38.700} & \underline{0.9385} & 0.0992 \\
    
    \bottomrule
    \end{tabular}
    }
    \caption{Quantitative denoising results on the real-world test set. “*” denotes the variant that applies SPAD-DSC before network training.}
    \label{tab:noise_model_comparison}
\end{table}

Tab.~\ref{tab:noise_model_comparison} and Fig.~\ref{fig:real_world_comparison} further evaluate whether the synthesized noise distributions lead to effective denoising on real SPAD images. Applying SPAD-DSC alone already outperforms the comparison baselines, showing that systematic sensor noise accounts for a large portion of the degradation in real SPAD images. This also explains why direct end-to-end training on noisy/clean pairs is insufficient. Without explicit correction, the network must learn fixed sensor artifacts together with shot noise, making systematic noise difficult to remove. Ours* first applies SPAD-DSC to remove most systematic noise and then uses the denoising network to suppress residual shot noise, achieving the best denoising performance. More results are provided in the supplementary material.
\subsection{Denoising Network Comparison}
Besides the U-Net baseline, we evaluate three Transformer-based denoisers, including Uformer~\cite{UFormer}, Restormer~\cite{Restormer}, and the Adaptive Sparse Transformer (AST~\cite{AST}) backbone used in our implementation. Replacing U-Net with stronger Transformer backbones improves denoising performance. However, when comparing each direct end-to-end variant (Ours) with its SPAD-DSC counterpart (Ours*), the improvement from SPAD-DSC is systematic across all backbones. This supports the accuracy of the proposed noise model and indicates that reliable physical priors are more effective than forcing the network to implicitly learn all sensor artifacts.

\subsection{Ablation Study}
We ablate the three calibrated quantities associated with the forward model in Eq.~\ref{eq:full-forward-model}: the pure dark-count rate $D_k$, the exposure-independent dark-noise bias $D_b$, and the correction gain $G=1/\eta$ corresponding to the relative pixel response $\eta$. When removing $D_k$ or $D_b$, we recalibrate the remaining dark-noise parameter under the corresponding reduced model rather than directly setting the removed term to zero.

Tab.~\ref{tab:noise_synthesis_ablation} reports the dark-frame synthesis ablation of $D_k$ and $D_b$; $G$ is excluded because it does not affect dark frames. Fig.~\ref{fig:noise_synthesis} further visualizes the same dark-noise ablations and their effect on noisy-image synthesis. Without $D_k$, the recalibrated $D_b$ represents an aggregate trigger level weighted across the calibration exposures and cannot describe exposure-dependent changes, leading to a large $R^2$ drop and visible distribution mismatch. Without $D_b$, the recalibrated $D_k$ uses a single slope to fit all latent dark-noise effects. This reduced model remains close to the full model but yields lower $R^2$ in three of the four dark-frame settings, showing that the additional bias term improves synthesis consistency across imaging configurations.

\begin{table}[h]
    \centering
    \resizebox{\columnwidth}{!}{
    \begin{tabular}{lcccccc}
    \toprule
    \multirow{2}{*}{\textbf{Ablation Setting}}
    & \multicolumn{3}{c}{\textbf{RAW Domain}}
    & \multicolumn{3}{c}{\textbf{sRGB Domain}} \\
    \cmidrule(lr){2-4} \cmidrule(lr){5-7}
    & \textbf{PSNR$\uparrow$} & \textbf{SSIM$\uparrow$} & \textbf{LPIPS$\downarrow$}
    & \textbf{PSNR$\uparrow$} & \textbf{SSIM$\uparrow$} & \textbf{LPIPS$\downarrow$} \\
    \midrule

    Ours* w/o $G$
    & 39.970 & 0.9911 & 0.0200 & 33.439 & 0.9248 & 0.0963 \\

    Ours* w/o $D_k$
    & 45.592 & 0.9907 & \underline{0.0183} & 36.434 & 0.9220 & \textbf{0.0835} \\

    Ours* w/o $D_b$
    & \underline{46.749} & \underline{0.9950} & \textbf{0.0152} & \underline{38.234} & \underline{0.9268} & \underline{0.0879} \\

    \midrule

    Ours*
    & \textbf{46.812} & \textbf{0.9952} & \textbf{0.0152} & \textbf{38.319} & \textbf{0.9274} & 0.0882 \\

    \bottomrule
    \end{tabular}
    }
    \caption{Ablation study of noise components.}
    \label{tab:ablation}
\end{table}

In the denoising ablation in Tab.~\ref{tab:ablation}, removing the gain correction causes the largest degradation, confirming that spatial response non-uniformity is a dominant systematic artifact in real SPAD images. Removing $D_k$ also clearly reduces performance, consistent with the dark-frame synthesis results. Removing $D_b$ has a smaller average effect, but the full model still achieves the best RAW/sRGB PSNR and SSIM, indicating that $G$, $D_k$, and $D_b$ play complementary roles in denoising.

\subsection{Real-world Applications}
We further demonstrate the practical value of SPAD intensity imaging in two challenging real-world scenarios: extremely low-light imaging and high-dynamic-range imaging. These examples are intended to show how the SPAD sensor characteristics and the proposed denoising pipeline translate to practical image acquisition. The CMOS comparison is captured with a Sony IMX174 sensor. Due to environmental capture constraints, the CMOS and SPAD views are not perfectly aligned, but the zoom-in regions are selected from approximately corresponding scene areas.

\begin{figure}[h]
    \centering
    \includegraphics[width=\columnwidth]{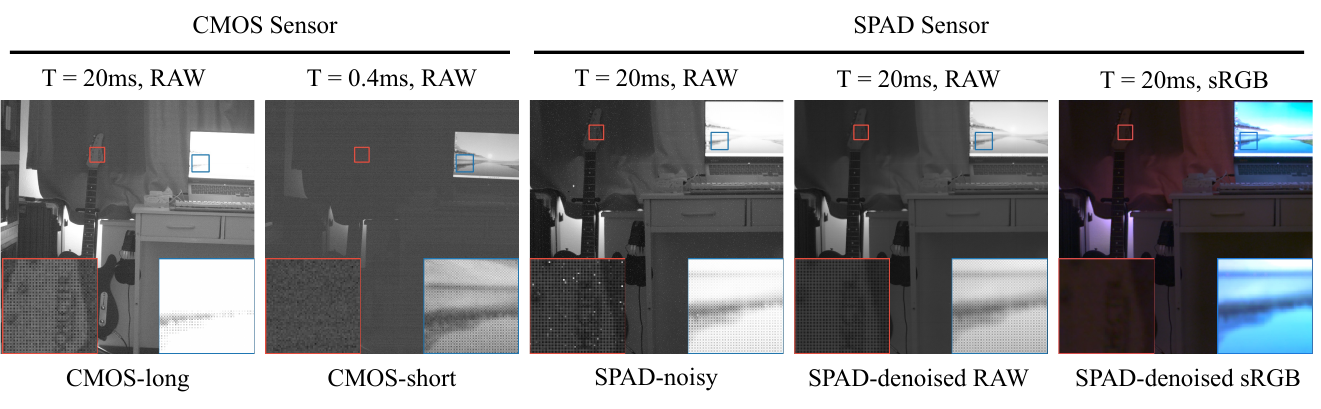}
    \caption{HDR imaging comparison between CMOS and SPAD cameras. Our method reduces SPAD noise while preserving both shadow structures and highlight details.}
    \label{fig:cvs_hdr}
\end{figure}

\begin{figure}[h]
    \centering
    \includegraphics[width=\columnwidth]{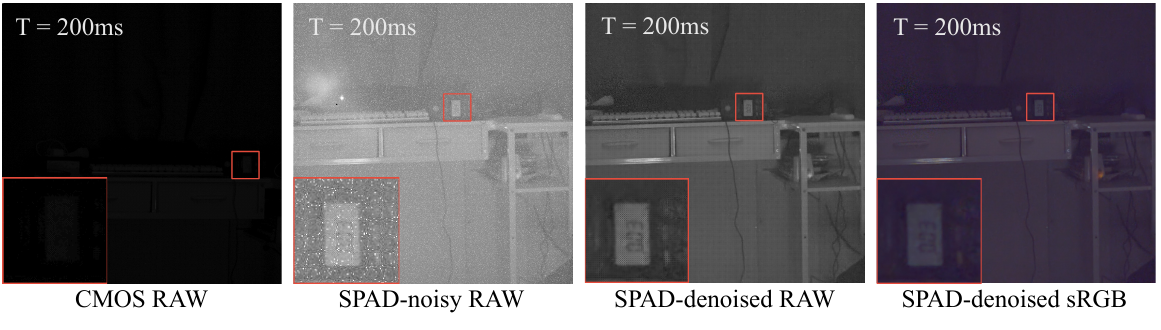}
    \caption{Low-light imaging comparison between CMOS and SPAD cameras. Our method suppresses strong SPAD noise while preserving visible scene structures.}
    \label{fig:cvs_lowlight}
\end{figure}

Fig.~\ref{fig:cvs_lowlight} shows a low-light scene (0.03 lx) captured with the same 200 ms exposure. The CMOS RAW image is severely underexposed, whereas the SPAD measurement already records recognizable content, though corrupted by strong sensor noise; our pipeline suppresses this noise while preserving the visible structures. Fig.~\ref{fig:cvs_hdr} shows an HDR scene where the CMOS sensor either saturates the bright display (long exposure) or loses the dark region (short exposure), while a single-exposure SPAD image retains both. Our denoising result removes the dominant noise while preserving the wide-dynamic-range content, illustrating the potential of SPAD intensity cameras for low-light and HDR photography.
\section{Conclusion}
We presented a practical noise modeling, calibration, and denoising framework for SPAD intensity imaging. Instead of explicitly decomposing ambiguous event sources, our model captures the observable dark-noise statistics with calibrated $D_k$ and $D_b$ terms and spatial non-uniformity with a relative response map, based on which we build a noise synthesis pipeline and a SPAD-specific dark-shading correction for network training. The framework is practical and effective in real-world SPAD denoising, even in the extreme low-light and HDR scenarios, demonstrating the potential of SPAD intensity imaging.

\subsubsection{Limitation.}
Limited by the available device, our noise model is built upon statistical observations of accumulated measurements. SPAD noise is also affected by factors such as temperature and timing jitter, and a SPAD sensor that precisely records photon arrival timestamps would facilitate modeling these effects. This work assumes a stable imaging environment without explicitly accounting for these effects. We leave these factors to future research.

\bibliography{aaai2027}

\clearpage
\def\supplementarytitle{Practical Noise Modeling for SPAD Intensity Imaging \\ Supplementary Material}
\makeatletter
\twocolumn[
  \vbox to \titlebox{%
    \hsize\textwidth%
    \linewidth\hsize%
    \vskip 0.625in minus 0.125in%
    \centering%
    {\LARGE\bf \supplementarytitle\par}%
    \vskip 1em plus 2fil%
  }%
]
\makeatother

\raggedbottom

\setcounter{section}{0}
\setcounter{subsection}{0}
\setcounter{figure}{0}
\setcounter{table}{0}
\setcounter{equation}{0}
\setcounter{secnumdepth}{2}
\renewcommand{\thesection}{\Alph{section}}
\renewcommand{\thesubsection}{\thesection.\arabic{subsection}}
\renewcommand{\thefigure}{S\arabic{figure}}
\renewcommand{\thetable}{S\arabic{table}}
\renewcommand{\theequation}{S\arabic{equation}}

\section{Calibration Details}

\subsection{Bad Pixel Detection}
We first repeat the dark-frame calibration used in the main paper. With incident light fully blocked, the observed binary triggers are sensor-induced. For each pixel and each exposure setting $t_i$, we count the number of triggered binary frames $k_i$ from $N_i$ valid observations and model the dark trigger statistics as
\begin{equation}
    k_i \sim \mathcal B\left(
    N_i,\; 1-\exp\left[-(D_k t_i + D_b)\right]
    \right),
    \label{eq:supp_dkdb_mle}
\end{equation}
where $D_k$ denotes the pure dark-count rate and $D_b$ denotes the exposure-independent dark-frame bias. The two parameters are estimated by constrained Binomial maximum likelihood estimation:
\begin{equation}
    (\hat D_k,\hat D_b)
    =\arg\min_{\substack{D_k\ge 0\\ D_b\ge 0}}
    \mathcal{L}(D_k,D_b),
    \label{eq:supp_dkdb_objective}
\end{equation}
\begin{equation}
    \begin{aligned}
    \mathcal{L}(D_k,D_b)
    =\sum_i\Big[
    &-k_i \log\!\big(1-\exp[-(D_k t_i+D_b)]\big) \\
    &+(N_i-k_i)(D_k t_i+D_b)
    \Big].
    \end{aligned}
    \label{eq:supp_dkdb_loss}
\end{equation}
Based on this multi-exposure dark calibration and the flat-field calibration, we diagnose four types of bad pixels:
\begin{itemize}
    \item \textbf{Hot pixels.} A pixel is marked as hot if its detection probability satisfies $p>0.5$ under any exposure setting. Such pixels trigger frequently even when incident light is fully blocked.
    \item \textbf{High-intercept pixels.} A pixel is marked as high-intercept if its fitted $D_b$ is above the $8\sigma$ upper threshold of the reference distribution. These pixels have high trigger levels that are largely independent of the gate time, which may be caused by crosstalk from nearby high-trigger pixels or afterpulsing effects.
    \item \textbf{Fitting outliers.} A pixel is marked as a fitting outlier if the $p$-value of the Pearson goodness-of-fit test is below $10^{-6}$ and its reduced Pearson statistic exceeds the $8\sigma$ upper threshold. We also include pixels whose trigger probability does not monotonically increase with exposure time. These pixels may be caused by random telegraph signal (RTS) noise or strong temperature-drift sensitivity, and therefore exhibit unstable statistics.
    \item \textbf{Dead pixels.} A pixel is marked as dead if its flat-field response is lower than $20\%$ of the local neighborhood response under the flat-field calibration condition.
\end{itemize}
These bad pixels make the dark-frame noise deviate from a stationary Poisson flow and produce obvious pepper-like noise in the accumulated SPAD observations. In real images, these pixels can overwhelm the signal from normal pixels. In our experiments, the above criteria detect $32623$ bad pixels, accounting for $3.1\%$ of the $1024\times1024$ sensor pixels. In SPAD-DSC, we replace each bad pixel with the mean of non-bad pixels in its local $3\times3$ neighborhood.
\begin{figure}[h]
  \centering
  \includegraphics[width=\columnwidth]{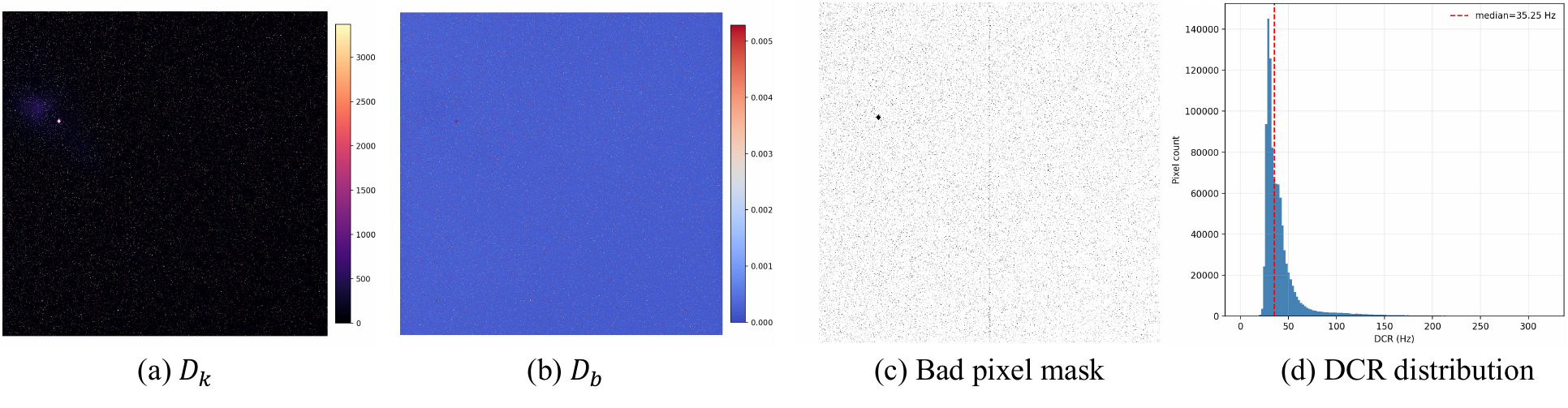}
  \caption{Dark-frame calibration results. The figure shows the calibrated $D_k$ map, the calibrated $D_b$ map, the detected bad-pixel mask, and the DCR distribution.}
  \label{fig:supp_dcr}
\end{figure}

\subsection{Flat-field Calibration}
\begin{figure*}[t]
  \centering
  \includegraphics[width=\textwidth]{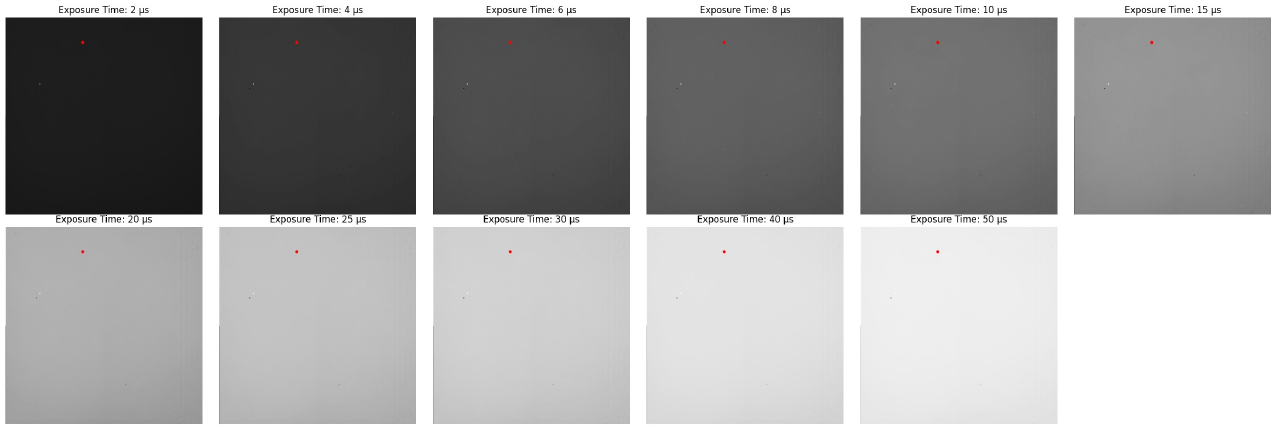}
  \caption{Flat-field frames captured under uniform illumination with exposure times from $2~\mu s$ to $50~\mu s$. The global intensity changes with exposure time, while the spatial shading pattern remains stable.}
  \label{fig:supp_flat}
\end{figure*}

Flat-field calibration estimates the spatial response non-uniformity of the SPAD sensor under uniform illumination. We capture 1-bit burst sequences with exposure times from $2~\mu s$ to $50~\mu s$ and accumulate them into different bit-depth settings. As shown in Fig.~\ref{fig:supp_flat}, the flat-field measurements become brighter as the exposure time increases, while the spatial shading pattern remains stable. This stable pattern indicates pixel-wise response non-uniformity caused by different sensitivities across the sensor, or lens vignetting effects.

\begin{figure*}[t]
  \centering
  \includegraphics[width=\textwidth]{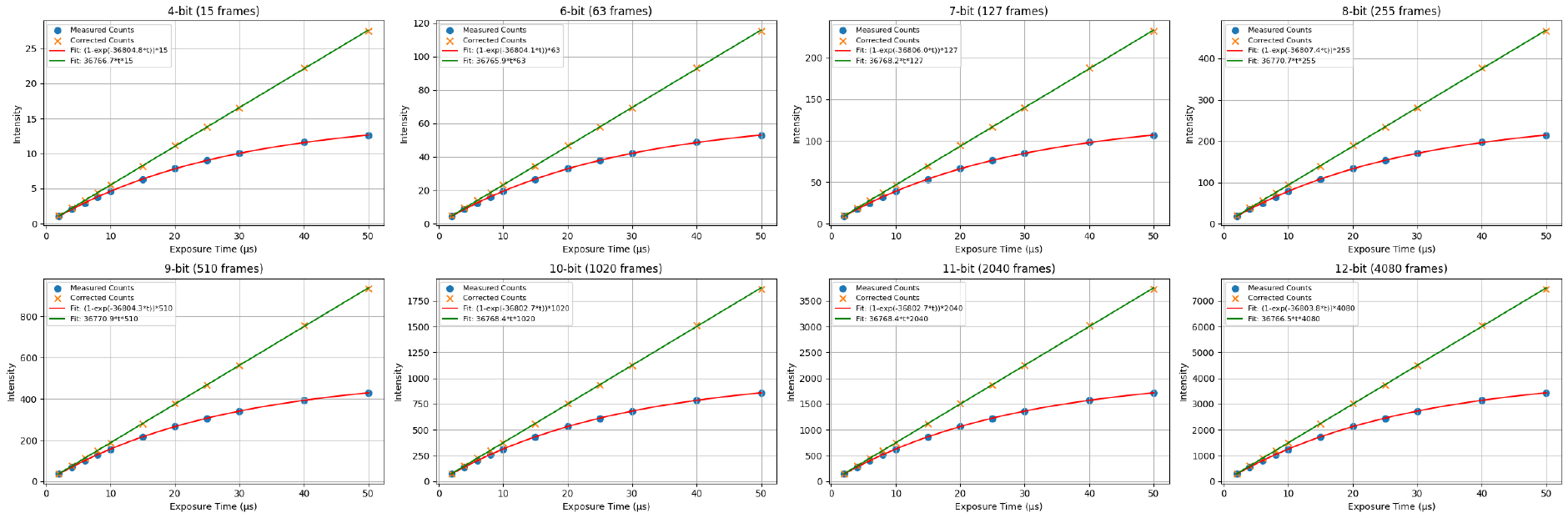}
  \caption{Pile-up correction for flat-field response calibration. Measured counts follow the nonlinear SPAD accumulation response, whereas the corrected counts become approximately linear with exposure time across different bit-depth settings.}
  \label{fig:supp_nonlinear}
\end{figure*}

Because SPAD intensity imaging records binary triggers, the accumulated counts follow a nonlinear pile-up response when the trigger probability increases. We therefore apply the same pile-up inversion used in the main paper before estimating the flat-field response. Fig.~\ref{fig:supp_nonlinear} verifies this correction across multiple accumulation depths. The measured counts follow the nonlinear SPAD response, while the pile-up-corrected counts become nearly linear with exposure time. This linearized response is then used to estimate the relative gain map.

\begin{figure}[t]
  \centering
  \includegraphics[width=\columnwidth]{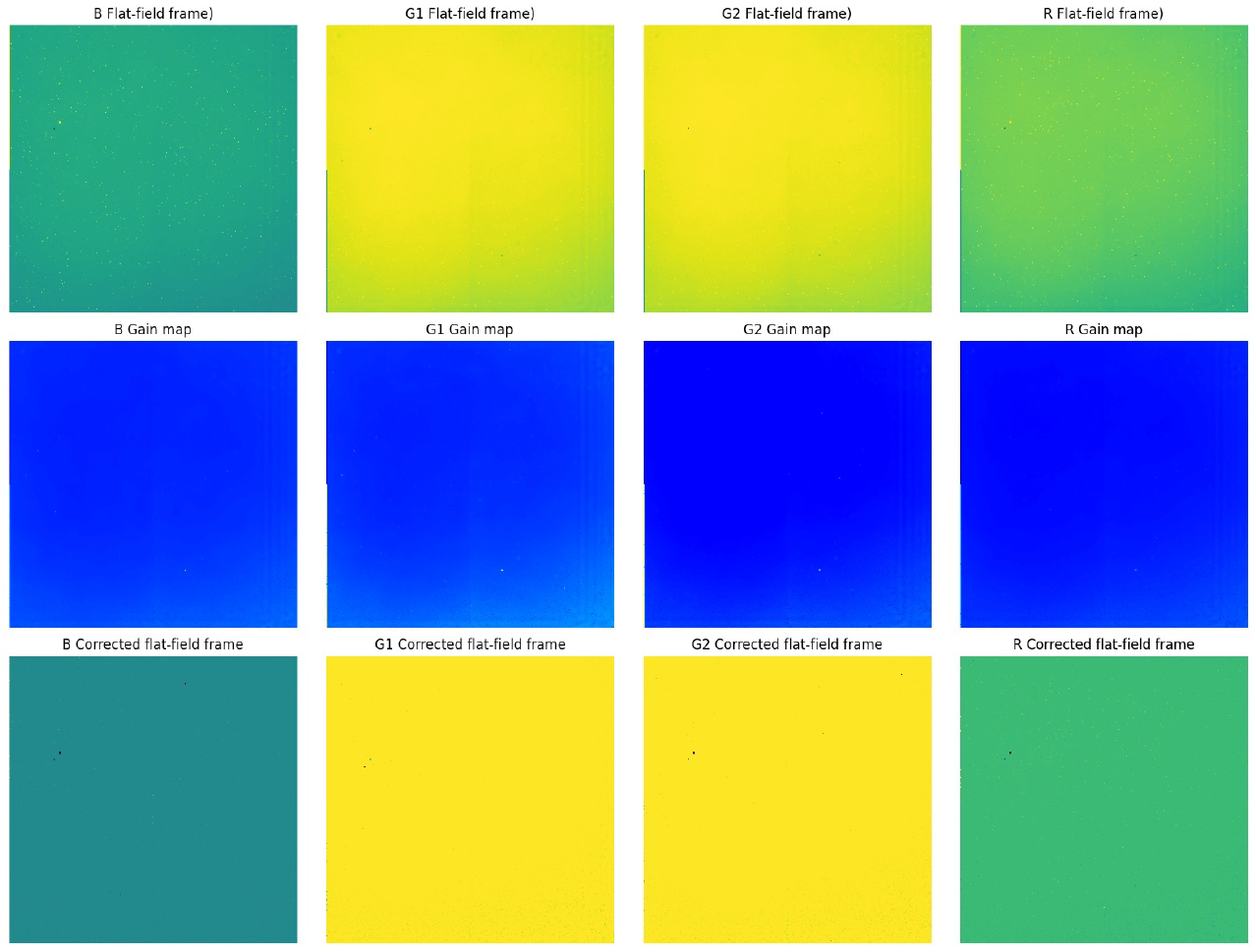}
  \caption{Channel-wise flat-field gain calibration. The top row shows the original flat-field frames, the middle row shows the estimated gain maps, and the bottom row shows the corrected flat-field frames for the B, G1, G2, and R channels.}
  \label{fig:supp_gain}
\end{figure}

After dark-noise subtraction and pile-up inversion, we compute the gain map separately for each Bayer channel. Fig.~\ref{fig:supp_gain} shows the original flat-field frames, the estimated channel-wise gain maps, and the corrected flat-field frames. The corrected frames exhibit substantially reduced spatial shading, indicating that the calibrated gain maps effectively compensate for pixel-wise and channel-wise response non-uniformity.


\subsection{Color ISP}
For color visualization and sRGB-domain evaluation, we use a calibrated lightweight color ISP instead of the camera internal color mode. The ColorChecker calibration image is captured in the 12-bit mode. During calibration, we enable the camera internal pile-up correction and noise correction, average about 100 frames, and then apply the calibrated non-uniformity correction and bad-pixel correction to obtain a stable RAW ColorChecker image.

The SPAD RAW image follows the BGGR Bayer pattern and is packed as four channels $[B,G1,G2,R]$ for processing. White-balance gains are estimated from the neutral gray patches of the ColorChecker, using the green channels as the reference. After applying WB gains, we demosaic the image with the BGGR Bayer pattern. To estimate the color correction matrix (CCM), we convert the standard ColorChecker sRGB values to the linear RGB domain using inverse sRGB gamma, extract the mean RGB value of each valid measured patch, and solve a least-squares color mapping:
\begin{equation}
    \mathbf c_{\mathrm{out}}
    = \mathbf c_{\mathrm{camera}}\mathbf M^\mathsf{T},
\end{equation}
where $\mathbf c_{\mathrm{camera}}$ denotes the measured linear RGB color and $\mathbf c_{\mathrm{out}}$ denotes the corrected linear RGB color. After CCM correction, the RGB values are clipped to $[0,1]$ and converted to sRGB with standard gamma correction.

\begin{figure*}[t]
  \centering
  \includegraphics[width=\textwidth]{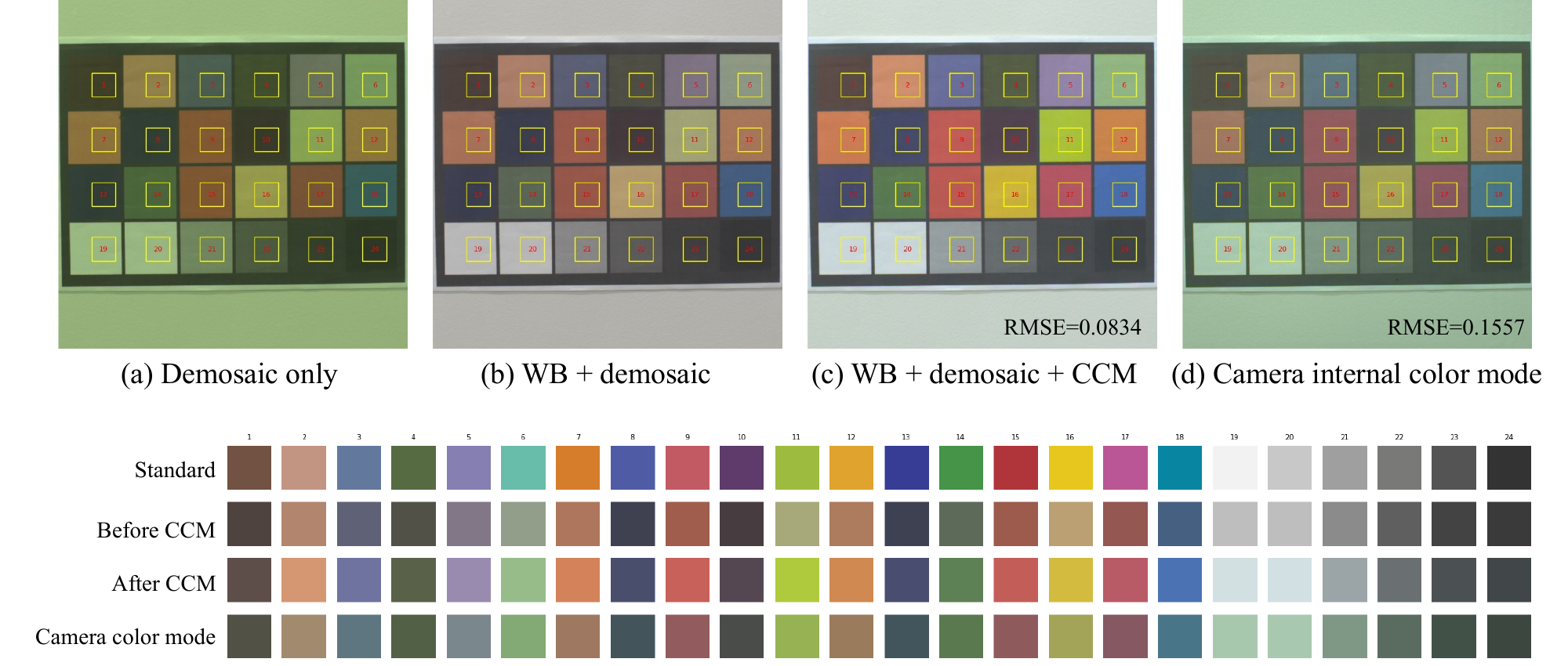}
  \caption{Color ISP calibration with a ColorChecker. The top row compares demosaicing only, WB with demosaicing, WB with demosaicing and CCM, and the camera internal color mode. The bottom row compares the standard ColorChecker colors with the measured patch colors before CCM, after CCM, and under the camera internal color mode.}
  \label{fig:supp_cisp}
\end{figure*}

Fig.~\ref{fig:supp_cisp} shows the effect of each color-processing step. Demosaicing alone produces a strong color bias, and WB reduces the neutral-color bias but still leaves clear color errors. After applying the calibrated CCM, the ColorChecker patches are much closer to the standard colors, reducing the patch RMSE to $0.0834$. In contrast, the camera internal color mode obtains a larger RMSE of $0.1557$ under the same ColorChecker evaluation. Therefore, all sRGB visualizations and evaluations in the paper use our calibrated WB and CCM pipeline.

\section{Implementation Details}

\subsection{Comparison Method Configuration}
Tab.~\ref{tab:supp_noise_formulation} summarizes the noise-synthesis formulations used for the comparison methods. Here $S$ denotes the clean SPAD-domain signal, $N$ is the number of accumulated binary frames, $\Delta t$ is the per-frame exposure time, and $D_{\rm real}$ denotes a matched real accumulated dark frame captured under the same bit-depth and exposure-time setting. For all baselines, the signal-dependent SPAD shot noise is modeled with the same binary-frame accumulation term so that the comparison focuses on the signal-independent noise model and its associated correction strategy.

\begin{center}
\begin{minipage}{\columnwidth}
\centering
\setlength{\tabcolsep}{0.08cm}
\resizebox{\columnwidth}{!}{
\begin{tabular}{ll}
\hline
\textbf{Model} & \textbf{Noise synthesis formulation} \\
\hline
Gaussian
&
$X=\mathcal B(N,1-e^{-S/N}) + \mathcal N(0,\sigma^2)$
\\
SFRN
&
$X=\mathcal B(N,1-e^{-S/N}) + D_{\rm real}$
\\
2-Shots
&
$X=\mathcal B(N,1-e^{-S/N}) + \widetilde D_{\rm 2shot}$
\\
Bian's model
&
$X\sim\mathcal B(N,q_{\rm Bian})$
\\
Ours
&
$X=\mathcal B(N,1 - e^{-\eta S/N - D_k\Delta t - D_b})$
\\
\hline
\end{tabular}
}
\captionof{table}{Noise synthesis formulations of different models.}
\label{tab:supp_noise_formulation}
\end{minipage}
\end{center}

\subsubsection{Gaussian Noise Model.}
The Gaussian baseline uses the SPAD binary-frame accumulation model for the signal term and adds independent Gaussian noise in the accumulated-count domain. The synthesized counts are clipped to the valid range $[0,N]$ before being used for training. This baseline represents a simple signal-independent and spatially unstructured noise model. Since it does not estimate SPAD-specific dark components or response non-uniformity, SPAD-DSC cannot be applied and the denoiser is trained directly on the synthesized noisy/clean pairs.

\subsubsection{SFRN.}
SFRN is adapted as an empirical real-dark-frame sampling baseline. For each bit-depth and exposure-time setting, we capture matched accumulated dark frames and randomly sample $D_{\rm real}$ from the corresponding dark-frame pool during training. This strategy preserves real dark-frame spatial structures and explains its strong dark-frame $R^2$ in the synthesis evaluation. However, the sampled accumulated dark frame is added to an independently sampled signal count image, rather than being combined with the signal at the binary-frame event level. SFRN also does not provide a SPAD-specific systematic correction model, so it is trained end-to-end without SPAD-DSC.

\subsubsection{Bian's Model.}
For Bian's model, we follow the original idea of explicitly decomposing binary dark-frame triggers into pure dark counts, afterpulsing, and crosstalk. For a binary dark-frame sequence $I_{\rm dark}(x,y,n)$, the dark-trigger formation model can be written as
\begin{equation}
\begin{aligned}
I_{\rm dark}(x,y,n)
=&\,N_{\rm dcr}(x,y,n)\\
&+p_{\rm ap}(x,y)I_{\rm dark}(x,y,n-1)\\
&+p_{\rm xt}(x,y)U(I_{\rm dark}(x,y,n)),
\end{aligned}
\label{eq:bian_model}
\end{equation}
where $N_{\rm dcr}$ denotes the pure dark-count component, $p_{\rm ap}$ denotes the afterpulsing probability map, $p_{\rm xt}$ denotes the crosstalk probability map, and $U(\cdot)$ summarizes the neighboring binary triggers around pixel $(x,y)$.

Bian et al. estimate these components by first assigning each dark-frame trigger to one of three categories:
\begin{itemize}
  \item A trigger is labeled as afterpulsing if the same pixel was triggered in the previous frame.
  \item Otherwise, it is labeled as crosstalk if a neighboring pixel is triggered in the same frame.
  \item The remaining triggers are labeled as pure dark counts.
  \item If both afterpulsing and crosstalk criteria are satisfied, the trigger is assigned to afterpulsing.
\end{itemize}
Let $I_{\rm ap}$ and $I_{\rm xt}$ denote the binary masks of triggers assigned to afterpulsing and crosstalk, respectively. The number of triggers assigned to pure dark counts is
\begin{equation}
\begin{aligned}
C_{\rm dc}(x,y)
=\sum_n\!\big[
&I_{\rm dark}(x,y,n)-I_{\rm ap}(x,y,n)\\
&-I_{\rm xt}(x,y,n)
\big].
\end{aligned}
\end{equation}
The corresponding probability maps are estimated by
\begin{equation}
\begin{aligned}
p_{\rm ap}(x,y)
=&\frac{\sum_n I_{\rm ap}(x,y,n)}{C_{\rm dc}(x,y)},\\
p_{\rm xt}(x,y)
=&\frac{\sum_n I_{\rm xt}(x,y,n)}{C_{\rm dc}(x,y)}.
\end{aligned}
\end{equation}
The pure dark-count term is estimated as
\begin{equation}
N_{\rm dcr}(x,y)=\frac{C_{\rm dc}(x,y)}{N_{\rm frames}}.
\end{equation}
In our comparison, we recalibrate these parameters on our captured binary dark-frame sequences following the same assignment rules. The calibrated pure-DCR, afterpulsing, and crosstalk terms are then combined into a marginal trigger probability $q_{\rm Bian}$ for count synthesis. This implementation follows the same decomposition logic as the original method. Since the model does not estimate the $D_k$, $D_b$, and gain terms used by SPAD-DSC, we train it directly on the synthesized noisy/clean pairs without applying our correction.

\subsubsection{2-Shots.}
2-Shots is adapted from CMOS RAW denoising by using matched SPAD dark frames. Following the original method, we first decompose a real accumulated dark frame $D_{\rm real}$ into a smooth dark-shading component and a zero-mean stochastic residual:
\begin{equation}
\begin{aligned}
S_{\rm ds}&=\mathcal G_{\sigma} * D_{\rm real},\\
R'&=D_{\rm real}-S_{\rm ds},\\
\mu_R&=\mathbb E[R'],\quad R=R'-\mu_R .
\end{aligned}
\end{equation}
The zero-mean residual $R$ is then used by the original Fourier-domain spectral sampler, which preserves the magnitude spectrum of $\mathcal F\{R\}$ and matches the residual histogram. Denoting the generated residual after this procedure as $N_{\rm spec}$, the synthesized dark frame is reconstructed by adding back the smooth shading and channel-wise mean:
\begin{equation}
\widetilde D_{\rm 2shot}=N_{\rm spec}+S_{\rm ds}+\mu_R .
\end{equation}
The original 2-Shots model uses Poisson noise for the signal-dependent term. In our SPAD comparison, we replace this term with the SPAD binary-frame accumulation model and synthesize the noisy count image as
\begin{equation}
X_{\rm 2shot}
=\mathcal B(N,1-e^{-S/N})+\widetilde D_{\rm 2shot}.
\end{equation}
Following the original 2-Shots setting, we also apply its own dark-shading correction before network training:
\begin{equation}
X_{\rm 2shot}^{\rm DSC}=X_{\rm 2shot}-S_{\rm ds}.
\end{equation}
This correction is independent of our SPAD-DSC and does not use the calibrated $D_k$, $D_b$, or gain maps.

\subsection{Training Configuration}
All U-Net denoising comparisons use the same training recipe so that performance differences mainly reflect the training-pair construction rather than optimization settings. The network takes packed four-channel RAW patches of size $512\times512\times4$ as input and predicts the corresponding clean packed RAW target. We train for $300$ epochs with the Charbonnier loss and the AdamW optimizer. The initial learning rate is $10^{-4}$, the weight decay is $10^{-4}$, and a cosine annealing scheduler reduces the learning rate to $10^{-6}$. The batch size is $4$.

Tab.~\ref{tab:supp_training_pair} summarizes the input-target pairs used by each compared method. Let $Y$ denote the clean SPAD-domain target generated from the SID RAW image. Each training input is constructed from the same clean target $Y$ using the corresponding noise model. For 2-Shots, we follow its own dark-shading correction. For Ours*, the input is corrected by the proposed SPAD-DSC before being passed to the network.

\begin{center}
\begin{minipage}{\columnwidth}
\centering
\setlength{\tabcolsep}{0.08cm}
\resizebox{\columnwidth}{!}{
\begin{tabular}{lll}
\hline
\textbf{Method} & \textbf{Network input} & \textbf{Training target} \\
\hline
Gaussian
& $\mathcal B(N,1-e^{-Y/N})+\mathcal N(0,\sigma^2)$
& $Y$
\\
SFRN
& $\mathcal B(N,1-e^{-Y/N})+D_{\rm real}$
& $Y$
\\
Bian's model
& $\mathcal B(N,q_{\rm Bian}(Y))$
& $Y$
\\
2-Shots
& $\mathcal B(N,1-e^{-Y/N})+\widetilde D_{\rm 2shot}-S_{\rm ds}$
& $Y$
\\
Ours
& $\mathcal B(N,1-e^{-\eta Y/N-D_k\Delta t-D_b})$
& $Y$
\\
Ours*
& $G[-N\log(1-X/N)-D_kT-ND_b]$
& $Y$
\\
\hline
\end{tabular}
}
\captionof{table}{Training input-target pairs for different methods. For Ours*, $X\sim\mathcal B(N,1-e^{-\eta Y/N-D_k\Delta t-D_b})$.}
\label{tab:supp_training_pair}
\end{minipage}
\end{center}
\begin{center}
  \begin{minipage}{\columnwidth}
  \centering
  \setlength{\tabcolsep}{0.08cm}
  \resizebox{\columnwidth}{!}{
  \begin{tabular}{ll}
  \hline
  \textbf{Method} & \textbf{Prediction} \\
  \hline
  Gaussian
  & $f_{\rm Gaussian}(X_{\rm real})$
  \\
  SFRN
  & $f_{\rm SFRN}(X_{\rm real})$
  \\
  Bian's model
  & $f_{\rm Bian}(X_{\rm real})$
  \\
  2-Shots
  & $f_{\rm 2shot}(X_{\rm real}-S_{\rm ds})$
  \\
  Ours
  & $f_{\rm ours}(X_{\rm real})$
  \\
  Ours*
  & $f_{\rm ours*}\!\left(G[-N\log(1-X_{\rm real}/N)-D_kT-ND_b]\right)$
  \\
  SPAD-DSC only
  & $G[-N\log(1-X_{\rm real}/N)-D_kT-ND_b]$
  \\
  \hline
  \end{tabular}
  }
  \captionof{table}{Real-image predictions for different methods.}
  \label{tab:supp_real_input}
  \end{minipage}
  \end{center}

\begin{figure*}[t]
  \centering
  \includegraphics[width=\textwidth]{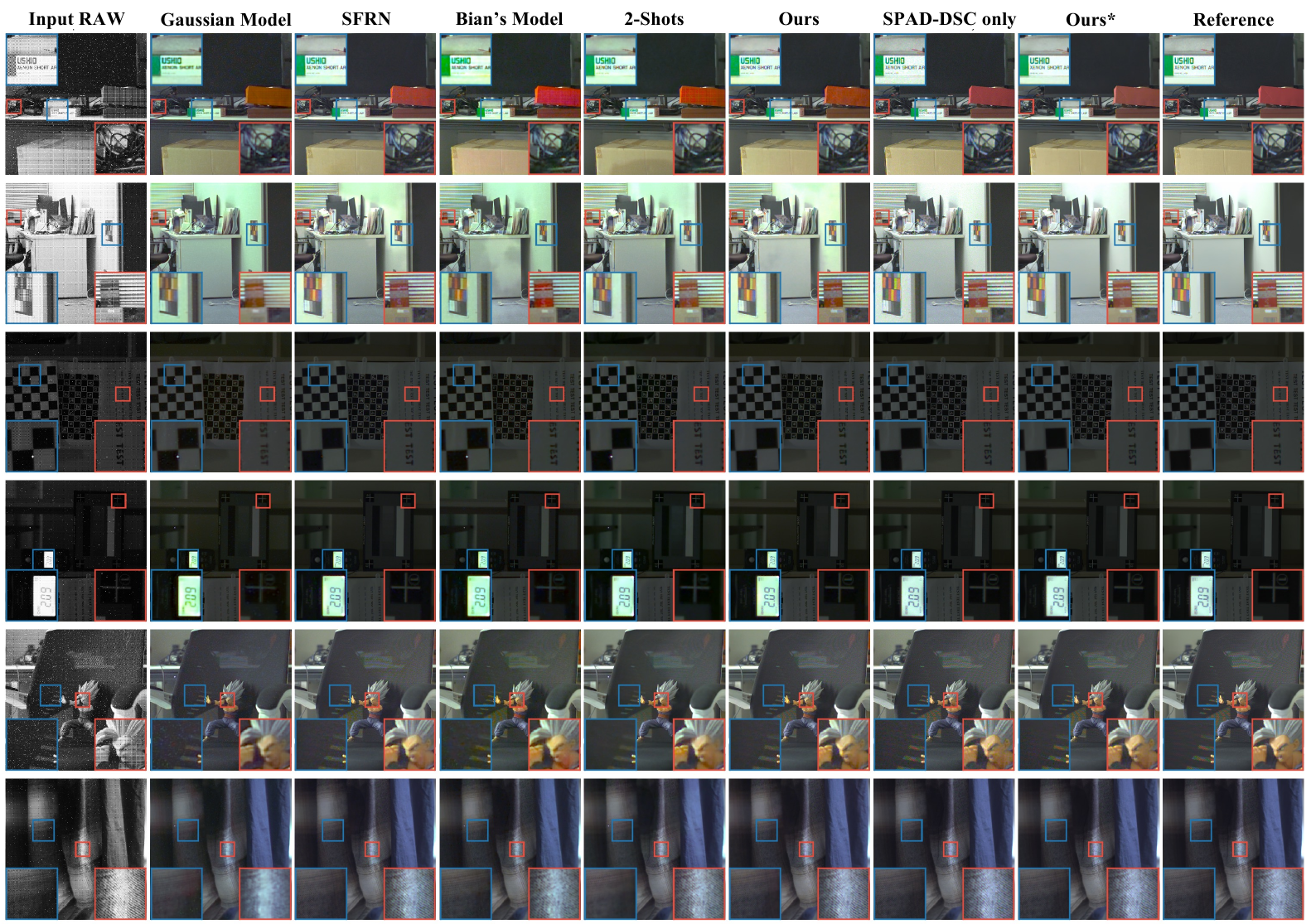}
  \caption{Additional qualitative denoising results on real-world scenes.}
  \label{fig:supp_qualitative}
\end{figure*}

During real-image evaluation, all methods are applied to the same noisy SPAD observation $X_{\rm real}$. Tab.~\ref{tab:supp_real_input} lists the prediction for each method, with the corresponding real-image input written inside the network function. We use $f_m$ to denote the denoising network trained with method $m$.

\section{Additional Experimental Results}
Fig.~\ref{fig:supp_qualitative} provides additional qualitative comparisons on real-world SPAD images. Ours* suppresses the structured dark noise and residual stochastic noise more effectively while preserving scene details.

\end{document}